\documentclass{article}
\usepackage[eabstract]{log_2026}						

\usepackage{booktabs}						
\usepackage{multirow}						
\usepackage{amsfonts}						
\usepackage{graphicx}						
\usepackage{float}	
\usepackage{enumitem}

\usepackage[numbers,sort]{natbib}	

\title[Diagnosing GNN Failure Modes with Layer-Wise Probe Entropy]
{Diagnosing GNN Failure Modes\\
with Layer-Wise Probe Entropy}

\author[S. Bao et al.]{%
Shanzi Bao\\
University of Cambridge\\
\email{sb2832@cl.cam.ac.uk}\And
Mar Gonzàlez I Català\\
University of Cambridge\\
\email{mg2211@cl.cam.ac.uk}\And
Pietro Liò\\
University of Cambridge\\
\email{pl219@cl.cam.ac.uk}
}

\newtheorem{definition}{Definition}

\begin{document}

\maketitle

\begin{abstract}

Graph neural networks (GNNs) are typically evaluated only at their final layer, obscuring how task-relevant information evolves during message passing. We introduce a post-hoc diagnostic that freezes a trained GNN, fits an independent linear probe to each layer, and tracks probe accuracy alongside normalized predictive entropy. Across real and synthetic graph settings, layer-wise probe entropy distinguishes beneficial, neutral, and harmful smoothing, a capability not demonstrated by existing geometric measures. Moreover, by comparing shallow and deep models, we show that the onset of entropy degradation separates the two main GNN failure modes: structural mismatch under heterophily, which appears from the earliest layers even in shallow models, and excessive-depth oversmoothing, which emerges only as propagation depth increases. These results establish probe-entropy trajectories as a tool for analyzing information flow in GNNs and motivate future work on adaptive depth selection and early-exit mechanisms.

\end{abstract}

\section{Introduction}
\label{sec:introduction}

Message-passing GNNs update node representations by aggregating information from neighbouring nodes~\cite{kipf2016semi,velivckovic2018graph}.
Across layers, task-relevant information may accumulate, remain stable, or degrade. Two distinct failure modes can cause this degradation. First, excessive propagation can oversmooth representations and erase discriminative information~\cite{li2018deeperinsightsgraphconvolutional,oono2020graph,nottoo, rusch2023survey}. Second, structural mismatch can arise under heterophily, where low-pass aggregation can mix incompatible label signals and become harmful from the first layers~\cite{gcnlowpass,zhu2020homophilygraphneuralnetworks,yan2022two}. A drop in final-layer accuracy can reveal that a failure has occurred, but it cannot distinguish between these cases. This distinction is important because the two failures require different remedies: oversmoothing may be addressed by limiting propagation depth, whereas structural mismatch may require changing the aggregation operator.

Existing approaches to identifying the source of failure have largely focused on detecting oversmoothing through changes in representation geometry~\cite{cosinesimilar,dsmooth1,dsmooth2,rusch2023survey}. These measures show that representations are becoming smoother, but changes in these smoothness measures alone do not establish whether smoothing improves, preserves, or degrades task performance. When smoothing is harmful, these measures alone also cannot determine whether the failure arises from excessive-depth oversmoothing or from aggregation that is harmful from the first layers under heterophily. 

Recent work on sequential reasoning uses predictive entropy across intermediate steps to track how answer-relevant information evolves during computation~\cite{català2026stepwiseinformativenessassumptionentropy, wang20258020rulehighentropyminority, agarwal2025unreasonableeffectivenessentropyminimization}. We study whether tracking entropy across GNN layers can similarly reveal how task-relevant information evolves and distinguish the two failure modes through their different trajectories: heterophily can be harmful from the first layers even in shallow models, whereas oversmoothing emerges only at greater total model depth. Concretely, we introduce a post-hoc diagnostic that tracks probe accuracy and predictive entropy across GNN layers and shows which of the two failure modes is occurring in controlled settings.

Our contributions are:
\begin{enumerate}
    \item We introduce a layer-wise probe entropy diagnostic to track how usable label information evolves during message passing.

    \item We demonstrate that probe entropy trajectories distinguish beneficial, neutral, and harmful smoothing, whereas Dirichlet energy and MAD do not consistently track these predictive consequences.

    \item We show that failure onset in probe entropy trajectories across shallow and deep models helps distinguish structural mismatch from oversmoothing in controlled settings across architectures.

    \item We analyze the reliability of the diagnostic across graph regimes, showing that per-node entropy-confidence alignment weakens under heterophily. 

\end{enumerate}

\section{Related Work}
\label{sec:background}

\paragraph{Failure-mode diagnostics in message-passing GNNs.}
Message-passing GNNs are subject to two main failure modes. The first is oversmoothing, a depth-related failure in which repeated message passing makes node representations increasingly similar and removes class-discriminative information~\cite{li2018deeperinsightsgraphconvolutional,oono2020graph,nottoo, rusch2023survey}. Oversmoothing is commonly diagnosed by tracking geometric quantities across layers: MAD measures distances between node representations~\cite{cosinesimilar}, while Dirichlet energy measures feature variation across graph edges~\cite{dsmooth1,dsmooth2}. These establish that smoothing occurs, but not its predictive consequence: whether it helps or harms task performance. Rank-based diagnostics track collapse more directly, with rank drops aligning with performance degradation~\cite{zhang2026we}. However, the dependence of rank on representation dimension complicates direct comparisons across layers. The group distance ratio $R_{\mathrm{Group}}$~\cite{zhou2020towards} instead compares inter-group to intra-group distances among node representations, but requires labels for the evaluated nodes. These diagnostics describe representation geometry and class separation. We complement them with layer-wise probe entropy, which tracks predictive uncertainty through a linear probe fitted at each layer.

The second failure mode arises under heterophily: low-pass aggregation can mix incompatible label signals and become harmful from the first message-passing layer, reflecting a structural mismatch between the aggregation operator and the task~\cite{gcnlowpass,zhu2020homophilygraphneuralnetworks}. Heterophily-aware architectures such as H2GCN and GPR-GNN~\cite{zhu2020homophilygraphneuralnetworks,chien2020adaptive} mitigate this mismatch but do not diagnose post hoc whether it has occurred in a trained model. A drop in final-layer accuracy can reveal that message passing has failed, but it cannot distinguish oversmoothing from structural mismatch~\cite{yan2022two}. We address this gap by comparing the onset of entropy degradation between a shallow and a deeper model, alongside probe accuracy, to help distinguish the two failure modes.

\paragraph{Layer-wise probing methods.}
Layer-wise probing provides a task-aware way to inspect intermediate representations. It freezes the model and fits an auxiliary classifier at each layer to measure how readily task labels can be decoded~\cite{alain2016understanding}. This approach has also been applied to graph models. For example,~\cite{pmlr-v206-akhondzadeh23a} uses probes at different layers to recover chemical and structural properties, while~\cite{pmlr-v284-pelletreau-duris25a} examines how graph-theoretic properties are represented across GNN layers. We instead use layer-wise probe entropy to assess whether smoothing improves, preserves, or degrades label readability, and to distinguish message-passing failure modes.

\section{Layer-Wise Probe Entropy}
\label{sec:signal}

After training, we freeze the GNN and fit an independent linear probe to each layer, including the input representation, using the labelled training nodes.

\begin{definition}[Normalized probe entropy]
Given a node \(v\) and layer \(k\), let \(h_v^{(k)}\) be the node representation,
\(
p^{(k)}(y \mid h_v^{(k)})
\)
be the probe's categorical distribution over the class label \(y\),
\(p_c^{(k)}(v)\) be the predicted probability that node \(v\) belongs to class
\(c\) and $C$ be the number of classes. Then \emph{normalized probe entropy} is defined as
\[
\widetilde{H}_k(v)
=
-\frac{1}{\log C}
\sum_{c=1}^{C}
p_c^{(k)}(v)\log p_c^{(k)}(v).
\] 
\end{definition}

For layer \(k\), we report the probe accuracy \(A_k\) and the mean entropy over the test-node set \(\mathcal{T}\),
\(
\overline{H}_k
=
\frac{1}{|\mathcal{T}|}
\sum_{v\in\mathcal{T}}
\widetilde{H}_k(v),
\)
where \(|\mathcal{T}|\) is the number of test nodes. Both quantities are summarized across independent runs.

We use probe accuracy to validate entropy as a diagnostic; once the probes are trained, entropy can be computed for any evaluation node without access to its label.

\section{Results}
\label{sec:results}

In this section, we evaluate whether layer-wise probe entropy can distinguish the predictive consequences of smoothing and the failure modes underlying harmful message passing. Our experiments span 12 real-world datasets, controlled homophily settings, multiple model depths, and both GCN and GAT. We seek to address the following questions:

\begin{enumerate}[
    label=(Q\arabic*),
    ref=Q\arabic*,
    leftmargin=2.8em,
    labelsep=0.8em,
    itemsep=0.4em,
    align=left
]
    \item\label{rq:smoothing}
    Can layer-wise probe entropy reveal whether smoothing is beneficial, neutral, or harmful?

    \item\label{rq:failure_mode}
    Can failure onset in probe entropy trajectories distinguish structural mismatch from excessive-depth oversmoothing?
\end{enumerate}

Complete experimental settings and dataset details are provided in Appendix~\ref{app:settings}. Additional experiments are presented in Appendix~\ref{app:additional_experiments}, with supporting analyses in Appendix~\ref{app:supporting_analyses}.

\subsection{Beneficial, Neutral, and Harmful Smoothing}
\label{sec:smoothing_consequences}

\begin{figure}[h]
    \centering
    \includegraphics[width=\textwidth]
    {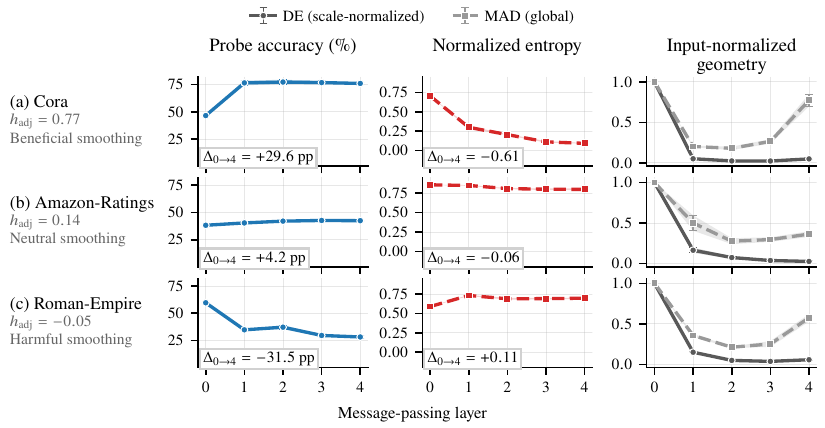}
   \caption{Beneficial, neutral, and harmful predictive outcomes under geometric smoothing in four-layer GCNs. Geometric trajectories are normalized by their input-layer values. Lines and markers show means across five seeds; shaded bands and error bars denote one standard deviation.}
    \label{fig:smoothing_consequences}
\end{figure}

Figure~\ref{fig:smoothing_consequences} answers \ref{rq:smoothing} by comparing probe entropy and geometric trajectories under beneficial,  neutral, and harmful smoothing.

\paragraph{(A1) Probe entropy trajectories distinguish beneficial,  neutral, and harmful smoothing.}
On Cora, message passing increases probe accuracy and lowers probe entropy. Amazon-Ratings, shown in the middle row of Figure~\ref{fig:smoothing_consequences}, exhibits a neutral response. On Roman-Empire, message passing decreases accuracy and raises entropy. Dirichlet energy decreases sharply in all three cases, while MAD changes substantially but non-uniformly. Therefore, neither Dirichlet energy nor MAD identifies whether smoothing is beneficial, neutral, or harmful.

We extend this comparison to 24 settings spanning 12 real-world datasets and GCN/GAT, including the label-aware group distance ratio \(R_{\mathrm{Group}}\)~\cite{zhou2020towards}, computed on training nodes using training labels only to match probe supervision. After averaging input-to-final-layer changes over five runs per setting, we correlate diagnostic changes with probe-accuracy changes. Table~\ref{tab:app_endpoint_correlations} shows that probe entropy has the largest observed absolute Pearson and Spearman correlations, with both 95\% dataset-cluster-bootstrap intervals excluding zero. For \(R_{\mathrm{Group}}\), only the Spearman interval excludes zero, whereas both intervals include zero for Dirichlet energy and MAD. Appendix~\ref{app:settings} defines the diagnostics. Appendix~\ref{app:a3_all_settings} details the protocol and presents the scatter plots in Figure~\ref{fig:a3_all_settings}.
\begin{table}[h]
    \centering
    \caption{Correlations between diagnostic changes and probe-accuracy changes across 24 dataset--architecture settings. Boldface indicates the largest absolute correlation for each coefficient.}
    \label{tab:app_endpoint_correlations}

    \setlength{\tabcolsep}{8pt}
    \renewcommand{\arraystretch}{1.15}

    \begin{tabular}{@{}l r@{}l c r@{}l c@{}}
        \toprule
        & \multicolumn{3}{c}{Pearson}
        & \multicolumn{3}{c}{Spearman} \\
        \cmidrule(lr){2-4}
        \cmidrule(lr){5-7}
        Diagnostic
        & \multicolumn{2}{c}{\(r\)} & 95\% CI
        & \multicolumn{2}{c}{\(\rho\)} & 95\% CI \\
        \midrule

        Probe entropy
        & \(\mathbf{-0}\) & \(\mathbf{.67}\)
        & \([-0.86,\,-0.34]\)
        & \(\mathbf{-0}\) & \(\mathbf{.71}\)
        & \([-0.89,\,-0.26]\) \\

        Dirichlet energy
        & \(-0\) & \(.07\)
        & \([-0.41,\,0.19]\)
        & \(-0\) & \(.27\)
        & \([-0.59,\,0.15]\) \\

        MAD
        & \(0\) & \(.01\)
        & \([-0.49,\,0.70]\)
        & \(0\) & \(.17\)
        & \([-0.43,\,0.66]\) \\

        \(R_{\mathrm{Group}}\)
        & \(0\) & \(.34\)
        & \([-0.09,\,0.72]\)
        & \(0\) & \(.62\)
        & \([0.03,\,0.82]\) \\

        \bottomrule
    \end{tabular}
\end{table}

\subsection{Failure-Onset Signatures Enable Failure-Mode Detection}
\label{sec:failure_mode_detection}

To answer \ref{rq:failure_mode}, we use GraphUniverse~\cite{graphuniverse} to generate controlled graphs at different homophily levels and independently train models at multiple total depths. Varying homophily controls structural mismatch between graph aggregation and label structure, while varying total depth allows us to study excessive-depth oversmoothing.

\begin{figure}[h]
    \centering
    \includegraphics[width=\linewidth]
    {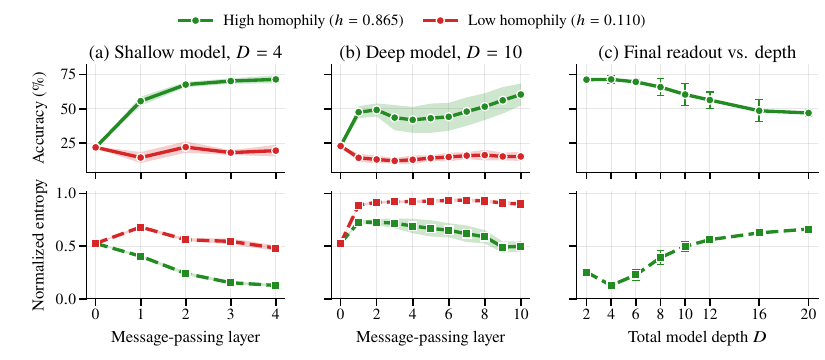}
\caption{Controlled GCN failure-onset trajectories. (a,b): Layer-wise probe accuracy and normalized entropy at
\(D=4,10\). (c): High-homophily final readouts across \(D\in\{2,4,6,8,10,12,16,20\}\).
Curves show means over five runs, with bands and error bars indicating one standard deviation.}
    \label{fig:failure_onset}
\end{figure}

\paragraph{(A2) Failure onset in probe entropy trajectories distinguishes
structural mismatch from oversmoothing.}
Failure onset refers to whether a harmful entropy increase appears immediately even in a shallow model or emerges only when the model becomes deep.
Figure~\ref{fig:failure_onset} compares these patterns. Under low homophily, entropy rises from the first message-passing layer in both the \(D=4\) and \(D=10\) models, accompanied by an accuracy drop. This immediate pattern across model depths indicates structural mismatch. Under high homophily, the shallow model improves accuracy and lowers entropy, whereas entropy rises from the first layer and remains high through most intermediate layers in the deep model. The depth grid also provides evidence of excessive-depth oversmoothing: beyond the shallow optimum, final accuracy decreases while final-layer entropy increases. The same patterns persist across eight model depths and both GCN and GAT. Complete controlled results and illustrative real-data trajectories on Cora and Roman-Empire are provided in Appendix~\ref{app:failure_onset_robustness}.

\section{Limitations}
\label{sec:scope}

Probe entropy measures the concentration of the predictive distribution rather than whether the prediction is correct. Confident predictions may therefore have low entropy even when they are wrong~\cite{guo2017calibrationmodernneuralnetworks, català2026stepwiseinformativenessassumptionentropy}. In our experiments, aggregate entropy trajectories track changes in probe accuracy, but node-level alignment between entropy and confidence in the true class weakens under heterophily. Here, alignment is measured by the per-node Spearman correlation across layers between probe entropy and the negative log-probability assigned to the true class. Table~\ref{tab:confidence_alignment} shows this pattern for both GCN and GAT at four and ten layers. Appendix~\ref{app:node_alignment} provides the full definition and controlled synthetic results, where alignment can approach zero or become negative under low homophily. Entropy-based diagnoses should therefore be validated in the target graph regime.

\begin{table}[h]
\centering
\caption{Per-node entropy--confidence alignment \(\overline{\rho}\). Values report mean \(\pm\) sample standard deviation across datasets within each regime, after averaging
five runs per dataset. Parentheses indicate dataset counts.}
\label{tab:confidence_alignment}
\setlength{\tabcolsep}{8pt}
\begin{tabular}{lcccc}
\toprule
& \multicolumn{2}{c}{GCN}
& \multicolumn{2}{c}{GAT} \\
\cmidrule(lr){2-3}
\cmidrule(lr){4-5}
Regime & 4 layers & 10 layers & 4 layers & 10 layers \\
\midrule
Homophilous (\(8\))
& \(0.63 \pm 0.22\) & \(0.44 \pm 0.20\)
& \(0.68 \pm 0.22\) & \(0.61 \pm 0.22\) \\
Heterophilous (\(4\))
& \(0.13 \pm 0.11\) & \(0.19 \pm 0.06\)
& \(0.12 \pm 0.13\) & \(0.16 \pm 0.08\) \\
\bottomrule
\end{tabular}
\end{table}

\section{Conclusion and Future Work}
\label{sec:conclusion}

We introduced layer-wise probe entropy as a post-hoc diagnostic of message passing. It distinguishes beneficial, neutral, and harmful smoothing, which Dirichlet energy and MAD alone do not reliably distinguish in our experiments. Comparing the onset of predictive degradation across shallow and deep models helps distinguish structural mismatch from excessive-depth oversmoothing in controlled settings.

As future work, the diagnostic should be evaluated across broader message-passing architectures and graph-learning tasks, as our experiments focus on GCN and GAT for node classification. Since our input, hidden, and output representations differ in dimension, we leave controlled comparisons with rank-based diagnostics at fixed representation dimension across model depths to future work.
A particularly exciting consequence of our findings is the potential for entropy trajectories to guide adaptive computation, for example by selecting the propagation depth at which entropy stops decreasing or enabling early exit once entropy is sufficiently low. Another direction is to train probes online alongside the GNN, using stop-gradient connections to prevent probe losses from updating the GNN, and assess their ability to track evolving representations during training.

\section*{Acknowledgements}
Shanzi Bao acknowledges travel support through the
G-Research NextGen Research Grant. Mar Gonzàlez I Català acknowledges that this project was supported by G-Research.

\bibliographystyle{unsrtnat}
\bibliography{reference}

\clearpage
\appendix
\section{Experimental Settings and Datasets}
\label{app:settings}

\paragraph{Datasets and splits.}
We evaluate 12 multiclass node-classification benchmarks spanning a broad range of graph structures. These include the Planetoid citation networks Cora, CiteSeer, and PubMed~\cite{yang2016revisiting}, the Amazon and Coauthor benchmarks~\cite{shchur2018pitfalls}, WikiCS~\cite{mernyei2020wiki}, and the heterophily benchmarks Amazon-Ratings, Chameleon, Squirrel, and Roman-Empire~\cite{platonov2023critical}. Adjusted homophily \(h_{\mathrm{adj}}\) is computed following~\cite{NEURIPS2023_01b68102}. For the controlled experiments, we generate graphs using GraphUniverse~\cite{graphuniverse}, with 500 nodes and 10 classes. The high-, intermediate-, and low-homophily conditions have measured edge homophily \(0.865\), \(0.512\), and \(0.110\), respectively, and use a fixed 60/20/20 train-validation-test split.

Real-dataset statistics and split protocols are summarized in Table~\ref{tab:datasets}. For Cora, CiteSeer, and PubMed, the class-balanced splits additionally contain 500 validation and 1,000 test nodes. For datasets with multiple splits, split \(i\) is paired with model run \(i\). Planetoid features are used without additional normalization, while Amazon and Coauthor features follow the standard benchmark preprocessing.

\begin{table}[h]
    \centering
    \caption{Real-world multiclass datasets and split protocols. For datasets with multiple splits, training-set sizes refer to the first split. \(h_{\mathrm{adj}}\) denotes adjusted homophily~\cite{NEURIPS2023_01b68102}.}
    \label{tab:datasets}

    \setlength{\tabcolsep}{4pt}
    \renewcommand{\arraystretch}{1.15}

    \begin{tabular*}{\linewidth}{
        @{\extracolsep{\fill}}lccccl@{}
    }
        \toprule
        Dataset
        & Nodes
        & Classes
        & Train nodes
        & \(h_{\mathrm{adj}}\)
        & Split protocol \\
        \midrule

        Coauthor-Physics
        & 34,493 & 5 & 20,695 & \(0.87\)
        & Fixed 60/20/20 \\

        Amazon-Photo
        & 7,650 & 8 & 4,590 & \(0.79\)
        & Fixed 60/20/20 \\

        Coauthor-CS
        & 18,333 & 15 & 10,999 & \(0.79\)
        & Fixed 60/20/20 \\

        Cora
        & 2,708 & 7 & 140 & \(0.77\)
        & 20 nodes per class \\

        PubMed
        & 19,717 & 3 & 60 & \(0.69\)
        & 20 nodes per class \\

        Amazon-Computer
        & 13,752 & 10 & 8,251 & \(0.68\)
        & Fixed 60/20/20 \\

        CiteSeer
        & 3,327 & 6 & 120 & \(0.67\)
        & 20 nodes per class \\

        WikiCS
        & 11,701 & 10 & 580 & \(0.58\)
        & Dataset-provided \\

        Amazon-Ratings
        & 24,492 & 5 & 12,246 & \(0.14\)
        & Official splits \\

        Chameleon
        & 890 & 5 & 409 & \(0.03\)
        & Filtered benchmark splits \\

        Squirrel
        & 2,223 & 5 & 1,053 & \(0.01\)
        & Filtered benchmark splits \\

        Roman-Empire
        & 22,662 & 18 & 11,331 & \(-0.05\)
        & Official splits \\

        \bottomrule
    \end{tabular*}
\end{table}

\paragraph{Models and training.}
To address Q1, we train four-layer GCNs and GATs on all 12 real-world datasets. To address Q2, we train both architectures on controlled graphs at depths \(D\in\{2,4,6,8,10,12,16,20\}\). We additionally use four- and ten-layer models for the node-level entropy--confidence analysis. GCN uses symmetrically normalized adjacency and 32-dimensional hidden representations. GAT uses four attention heads with eight dimensions per head, giving the same total hidden dimensionality. Both architectures use ReLU activations and are trained for 100 epochs with Adam and a learning rate of \(0.01\).

\paragraph{Linear probes.}
For each input and message-passing representation, we train the independent linear probe described in Section~\ref{sec:signal} for 100 epochs after freezing the GNN. Probes are fitted on the training nodes and evaluated on the test nodes.

\paragraph{Geometric diagnostics.}
We compute all three diagnostics on the same layer representations used by the probes. These are the input features at \(k=0\), post-activation hidden representations at \(1\leq k<K\), and the pre-softmax output at \(k=K\).

\subparagraph{Scale-normalized Dirichlet energy.}
For a layer-\(k\) representation matrix \(H^{(k)}\), we report Dirichlet energy in scale-normalized form, also known as the Dirichlet Rayleigh quotient:
\[
E_{\mathrm{SN}}(H^{(k)})
=
\frac{
\operatorname{tr}\!\left(
H^{(k)\top}\widetilde{\Delta}H^{(k)}
\right)
}{
\lVert H^{(k)}\rVert_F^2
}
=
\frac{
\frac12\sum_{i,j}\widetilde a_{ij}
\left\|
\frac{h_i^{(k)}}{\sqrt{\widetilde d_i}}
-
\frac{h_j^{(k)}}{\sqrt{\widetilde d_j}}
\right\|_2^2
}{
\lVert H^{(k)}\rVert_F^2
}.
\]
Here, \(\widetilde A=A+I\),
\(\widetilde d_i=\sum_j\widetilde a_{ij}\), and
\(\widetilde{\Delta}
=I-\widetilde D^{-1/2}\widetilde A\widetilde D^{-1/2}\)
is the augmented normalized Laplacian. We use this scale-normalized form rather than raw Dirichlet energy because raw energy scales quadratically with representation magnitude, and its layer-wise trajectory therefore conflates geometric smoothing with norm changes induced by the learned transformations~\cite{rank}. The Rayleigh quotient is also the quantity that~\cite{dsmooth1} identify as the real measure of smoothness, in contrast to the unnormalized energy tracked by their decay analysis.

\subparagraph{Global MAD.}
Following~\cite{cosinesimilar}, we compute the cosine distance between nodes \(i\) and \(j\) at layer \(k\) as
\[
D_{ij}^{(k)}
=
1-
\frac{
\langle h_i^{(k)},h_j^{(k)}\rangle
}{
\lVert h_i^{(k)}\rVert_2
\lVert h_j^{(k)}\rVert_2
}.
\]
We use global MAD, which averages this distance over all distinct node pairs:
\[
\mathrm{MAD}^{\mathrm{global}}(H^{(k)})
=
\frac{1}{n(n-1)}
\sum_{i=1}^{n}
\sum_{\substack{j=1\\j\neq i}}^{n}
D_{ij}^{(k)},
\]
where \(n\) is the number of nodes. Lower values indicate that node representations are more globally similar. We abbreviate \(\mathrm{MAD}^{\mathrm{global}}\) as MAD throughout.

\subparagraph{Group distance ratio.}
Following the implementation of~\cite{zhou2020towards}, we evaluate the label-aware group distance ratio, which compares inter-class and intra-class squared Euclidean distances between L2-normalized node representations.
Let \(V_{\mathrm{train}}\) denote the training-node set, \(y_v\) the class label of node \(v\), and \(h_v^{(k)}\) its representation at layer \(k\).
For \(C\) classes, define \(L_c=\{v\in V_{\mathrm{train}}:y_v=c\}\), with \(|L_c|\) denoting the number of training nodes in class \(c\).
We first normalize each node representation as
\[
\hat h_v^{(k)}
=
\frac{h_v^{(k)}}{
\max\!\left(\lVert h_v^{(k)}\rVert_2,\varepsilon\right)},
\qquad \varepsilon=10^{-12}.
\]
We then compute
\[
R_{\mathrm{Group}}^{(k)}
=
\frac{
\displaystyle
\frac{1}{C(C-1)}
\sum_{c=1}^{C}
\sum_{\substack{d=1\\d\neq c}}^{C}
\frac{1}{|L_c||L_d|}
\sum_{u\in L_c}\sum_{v\in L_d}
\left\lVert \hat h_u^{(k)}-\hat h_v^{(k)}\right\rVert_2^2
}{
\displaystyle
\frac{1}{C}
\sum_{c=1}^{C}
\frac{1}{|L_c|^2}
\sum_{u\in L_c}\sum_{v\in L_c}
\left\lVert \hat h_u^{(k)}-\hat h_v^{(k)}\right\rVert_2^2
}.
\]
Larger values indicate greater inter-class separation relative to within-class variation in the normalized representation space. We compute this diagnostic on the same layer representations as the other measures, using only training nodes and their labels to match the label access available to the probes.

\paragraph{Seeds and aggregation.}
All experiments use five model seeds \(\{42,123,456,789,1024\}\). Plots report the mean and standard deviation across seeds unless stated otherwise. Per-node entropy and confidence Spearman correlations are computed within each seed before being summarized across seeds.


\section{Additional Experiments}
\label{app:additional_experiments}

\subsection{Across-Setting Predictive and Geometric Changes}
\label{app:a3_all_settings}

We compare input-to-final-layer changes across 12 real-world datasets and two architectures, using four-layer models and five runs per setting.

For each run \(s\), we compute the changes in probe accuracy and predictive entropy,
\[
\Delta A_s=A_{4,s}-A_{0,s},
\qquad
\Delta\overline H_s=\overline H_{4,s}-\overline H_{0,s},
\]
and the relative changes in the three geometric diagnostics,
\[
\Delta E_s
=
\frac{E_{\mathrm{SN},s}^{(4)}}{E_{\mathrm{SN},s}^{(0)}}-1,
\quad
\Delta\mathrm{MAD}_s
=
\frac{\mathrm{MAD}_s^{(4)}}{\mathrm{MAD}_s^{(0)}}-1,
\quad
\Delta R_{\mathrm{Group},s}
=
\frac{R_{\mathrm{Group},s}^{(4)}}{R_{\mathrm{Group},s}^{(0)}}-1.
\]

Probe accuracy lies in \([0,1]\), and probe entropy is normalized to the same range by dividing by \(\log C\), so we report direct differences for both (accuracy changes in percentage points). For the geometric diagnostics, we use relative changes to compare proportional changes across settings with different input-layer baseline values.
We average each change across the five runs, producing one value per dataset--architecture setting, and compute Pearson and Spearman correlations across the resulting 24 setting means. For both coefficients, 95\% confidence intervals are obtained using 10,000 dataset-cluster-bootstrap draws. In each draw, we resample the 12 datasets with replacement, retain the paired GCN and GAT settings for each sampled dataset, and recompute both correlation coefficients.

\begin{figure}[h]
    \centering
    \includegraphics[width=\linewidth]
    {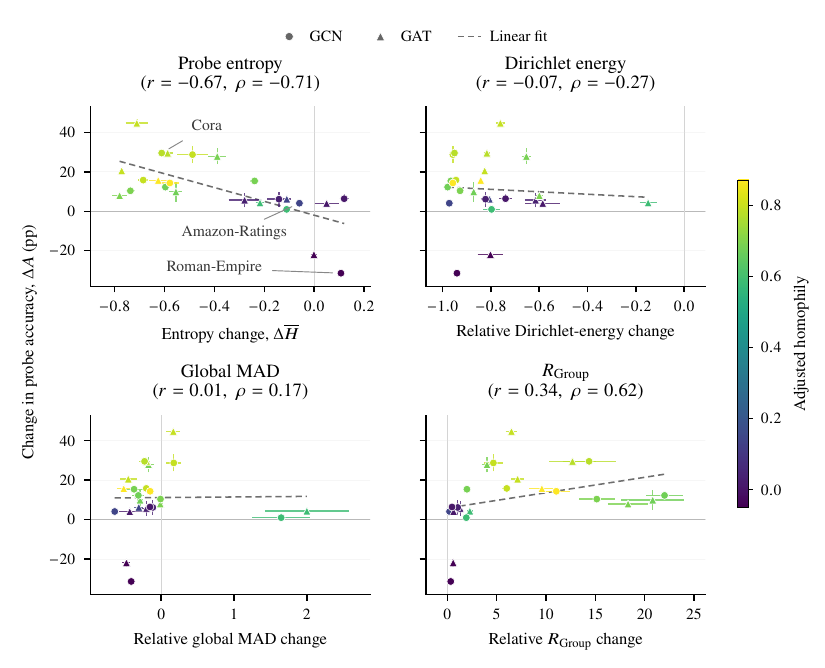}
    \caption{Input-to-final-layer probe-accuracy change versus
    changes in probe entropy, scale-normalized Dirichlet energy,
    global MAD, and \(R_{\mathrm{Group}}\) across 24 real
    dataset--architecture settings with four-layer models.
    Geometric changes are relative to input-layer values.
    Circles and triangles denote GCN and GAT, respectively,
    and colour indicates adjusted homophily.
    Points and error bars show the mean and standard deviation
    across five runs. Dashed lines show linear fits;
    panel titles report Pearson and Spearman correlations.}
    \label{fig:a3_all_settings}
\end{figure}

Figure~\ref{fig:a3_all_settings} compares the four diagnostics across settings. The labelled GCN settings correspond to the three representative datasets in Figure~\ref{fig:smoothing_consequences}. Probe entropy shows the largest observed absolute Pearson and Spearman correlations with accuracy change, as reported in Table~\ref{tab:app_endpoint_correlations}. Large reductions in Dirichlet energy occur alongside both gains and losses in accuracy. Changes in MAD also show weak associations with accuracy changes, whereas \(R_{\mathrm{Group}}\) better tracks their ordering, consistent with its positive Spearman correlation.

\paragraph{Sensitivity of DE and MAD to endpoint parameterization.}
Because Dirichlet energy and MAD can approach zero after message passing, their relative changes may concentrate near \(-1\), compressing differences across settings. We therefore average the endpoint ratios across the five runs within each setting and take the logarithm of the resulting 24 mean ratios. Using the same dataset-cluster bootstrap protocol, the Pearson associations remain weak for scale-normalized Dirichlet energy (\(r=-0.05\), 95\% CI \([-0.38,0.16]\)) and MAD (\(r=0.15\), 95\% CI \([-0.44,0.68]\)). Because the logarithm preserves the ordering of the setting-level values, the Spearman correlations and their confidence intervals remain unchanged (\(\rho=-0.27\) and \(0.17\), respectively). Thus, the observed associations for DE and MAD remain weak under the log-ratio parameterization.

\subsection{Failure-Onset Robustness Across Depths and Architectures}
\label{app:failure_onset_robustness}

This appendix tests whether the controlled failure-onset patterns generalize across model depths and architectures, and whether similar trajectories can be observed on real datasets. We use the controlled GraphUniverse graphs and experimental protocol described in Appendix~\ref{app:settings}, with five runs per setting.

\paragraph{Controlled robustness across depths and architectures.}
To validate the two failure patterns, we test whether first-step entropy increases under low homophily in both shallow and deep models, and whether under high homophily it decreases in shallow models but increases in deeper models. We also test whether the latter change is accompanied by higher final entropy and lower final accuracy as total model depth increases.

\begin{figure}[h]
    \centering
    \includegraphics[width=\linewidth]
    {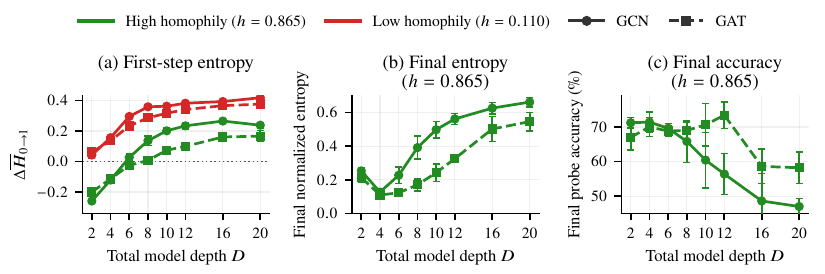}
    \caption{Controlled failure-onset patterns across eight depths for GCN and GAT. Panel (a) shows first-step entropy changes under high and low homophily. Panels (b)--(c) show final entropy and accuracy under high homophily. Error bars denote one standard deviation across five runs.}
    \label{fig:appendix_onset_summary}
\end{figure}

In panel (a), values above zero indicate that entropy increases after the first message-passing layer, while values below zero indicate that it decreases. The low-homophily curves remain above zero at every tested depth for both GCN and GAT, with this positive change holding in \(80/80\) runs. The high-homophily curves are below zero for all \(20/20\) shallow runs at \(D\in\{2,4\}\), but cross above zero as total model depth increases. Panels (b)--(c) show that, beyond the shallow optimum, this transition is accompanied by higher final entropy and lower final accuracy for both architectures.

\begin{figure}[h]
    \centering
    \includegraphics[width=\linewidth]
    {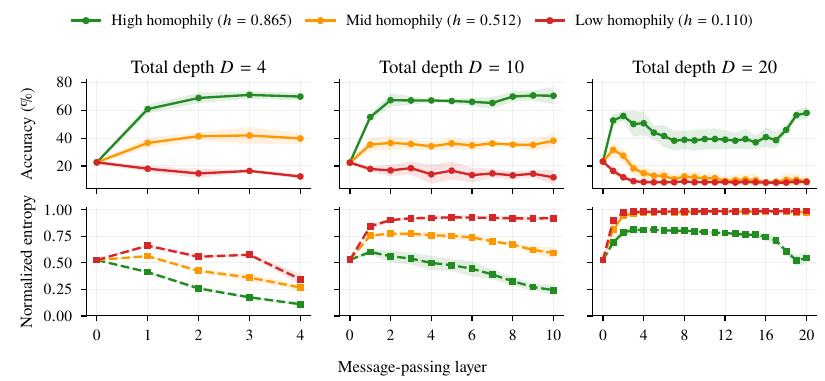}
    \caption{Controlled GAT failure-onset trajectories at \(D\in\{4,10,20\}\). Curves show mean \(\pm\) standard deviation across five runs.}
    \label{fig:appendix_gat_onset}
\end{figure}

\paragraph{Full GAT trajectories.}
{
Figure~\ref{fig:failure_onset} presents the complete layer-wise trajectories for GCN. As an architectural counterpart, Figure~\ref{fig:appendix_gat_onset} shows the corresponding GAT trajectories at representative shallow, intermediate, and deep configurations.
}

The GAT trajectories reproduce the immediate low-homophily pattern and the contrast between beneficial shallow propagation and high-entropy deep propagation under high homophily.

\paragraph{{Representative real-data trajectories.}}

\begin{figure}[H]
    \centering
    \includegraphics[width=\linewidth]
    {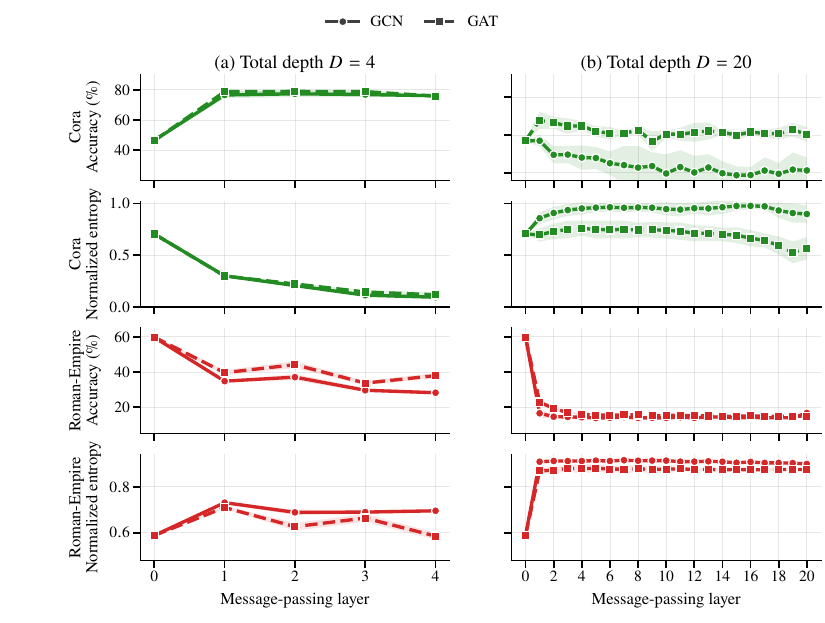}
    \caption{Representative layer-wise probe trajectories on Cora and Roman-Empire for shallow and deep GCNs and GATs. Lines and shaded bands show the mean and one standard deviation across five runs.}
    \label{fig:app_real_trajectories}
\end{figure}

Figure~\ref{fig:failure_onset} establishes the two trajectory patterns on controlled graphs. To examine whether qualitatively similar patterns also appear on real datasets, Figure~\ref{fig:app_real_trajectories} shows complete layer-wise trajectories on Cora and Roman-Empire.

On high-homophily Cora, shallow models progressively lower entropy, whereas deep models retain higher entropy and produce poorer accuracy. On low-homophily Roman-Empire, entropy rises and accuracy falls from the first message-passing layer in both shallow and deep models. These trajectories match the corresponding patterns observed in the controlled experiments.

\section{Supporting Analyses}
\label{app:supporting_analyses}

\subsection{Node-Level Entropy--Confidence Alignment}
\label{app:node_alignment}

The main results use entropy averaged over test nodes to diagnose population-level message-passing dynamics. However, applications such as node-level uncertainty estimation or early exiting require entropy to remain informative for individual nodes. We therefore examine whether changes in a node's entropy across layers are aligned with changes in the confidence assigned to its true class.

Let \(\mathcal{T}\) denote the set of test nodes, \(v\in\mathcal{T}\) a test node, and \(K\) the total number of message-passing layers. The index \(k=0\) denotes the input representation, while \(k=1,\ldots,K\) denotes the successive message-passing layers. For node \(v\), \(\widetilde H_k(v)\) is the normalised entropy of the layer-\(k\) probe distribution. We define the probability assigned to the true class \(y_v\) as

\[
{
p_k^*(v)
=
p^{(k)}\!\left(y_v\mid h_v^{(k)}\right),
}
\]

where \(h_v^{(k)}\) is the layer-\(k\) representation and \(p^{(k)}\) is the corresponding probe distribution. Thus, \(-\log p_k^*(v)\) increases as the probe becomes less confident in the true class.

For each node, we compute the Spearman correlation between its entropy and true-class negative log-probability trajectories:

\begin{equation}
    {
    \rho(v)
    =
    \operatorname{Spearman}
    \left(
        \{\widetilde H_k(v)\}_{k=0}^{K},
        \{-\log p_k^*(v)\}_{k=0}^{K}
    \right).
    }
\end{equation}

The correlation is computed over the \(K+1\) layer readouts for node \(v\). A positive value indicates that entropy increases when confidence in the true class decreases. Spearman correlation is useful here because it measures monotonic alignment without requiring a linear relationship or comparable numerical scales between the two trajectories.

We then average the node-level correlations over the test set:

\begin{equation}
    {
    \overline{\rho}
    =
    \frac{1}{|\mathcal{T}|}
    \sum_{v\in\mathcal{T}}
    \rho(v),
    }
\end{equation}

where \(|\mathcal{T}|\) is the number of test nodes. We first compute \(\overline{\rho}\) within each run and then summarise it across runs and datasets.

\begin{table}[h]
\centering
\caption{Per-node entropy--confidence Spearman correlation \(\overline{\rho}\) on controlled synthetic graphs. Values report mean \(\pm\) standard deviation across five seeds.}
\label{tab:synthetic_spearman}
\setlength{\tabcolsep}{6pt}
\begin{tabular}{lccccc}
\toprule
& & \multicolumn{2}{c}{GCN} & \multicolumn{2}{c}{GAT} \\
\cmidrule(lr){3-4}
\cmidrule(lr){5-6}
Regime
& Edge homophily
& 4 layers
& 10 layers
& 4 layers
& 10 layers \\
\midrule
High
& \(0.865\)
& \(0.50 \pm 0.03\)
& \(0.36 \pm 0.07\)
& \(0.51 \pm 0.04\)
& \(0.47 \pm 0.06\) \\

Intermediate
& \(0.512\)
& \(0.17 \pm 0.04\)
& \(0.13 \pm 0.04\)
& \(0.06 \pm 0.02\)
& \(0.14 \pm 0.04\) \\

Low
& \(0.110\)
& \(-0.10 \pm 0.03\)
& \(0.02 \pm 0.05\)
& \(-0.27 \pm 0.07\)
& \(0.02 \pm 0.05\) \\
\bottomrule
\end{tabular}
\end{table}

Table~\ref{tab:synthetic_spearman} shows that alignment on controlled synthetic graphs weakens as homophily decreases, becoming negative for four-layer models in the low homophily setting. These results complement the real-world benchmark results in Section~\ref{sec:scope} and highlight the need to validate node-level uses of probe entropy under heterophily.

\end{document}